# Exploring Language-Independent Emotional Acoustic Features via Feature Selection

Arslan Shaukat and Ke Chen, *Senior Member, IEEE*

*Abstract*—We propose a novel feature selection strategy to discover language-independent acoustic features that tend to be responsible for emotions regardless of languages, linguistics and other factors. Experimental results suggest that the language-independent feature subset discovered yields the performance comparable to the full feature set on various emotional speech corpora.

## I. Introduction

VARIOUS acoustic features for emotional speech have been investigated recently, which are based on both utterance-based and segment-based approaches for feature extraction [1-3]. It has been reported that the success in emotion recognition from speech is achieved by the combination of certain features [2]. Numerous feature selection techniques have been applied to acoustic features in order to find out those features responsible for emotional speech [1], [2], [4], [5], [6]. In their studies, however, feature selection is made on a specific corpus. In general, selected features are effective for a single corpus only and not generalized to other corpora in general.

Unlike the previous work, we aim at discovering those acoustic features that tend to be responsible for emotions regardless of linguistics and other factors and hence can be generalized to other corpora. In this paper, we name such features as *language-independent emotional acoustic features*. Human beings can generally recognize emotions from speech regardless of languages as they exploit some kind of acoustic features irrespective of linguistics [7]. Motivated by the human perceptual experience, we propose a novel feature selection strategy to explore language-independent emotional acoustic features regardless of linguistics, semantics and design of the emotional speech. In our work, feature selection techniques are applied to a training corpus, and selected features are then tested on corpora of different languages and designs. We name such a feature selection strategy . By using the  strategy, we discover language-

Authors are with the School of Computer Science, The University of Manchester, Kilburn Building, Oxford Road, Manchester M13 9PL, United Kingdom (phone: 44-161-3064565; fax: 44-161-2756204; e-mail: Arslan.Shaukat@postgrad.manchester.ac.uk, Chen@cs.manchester.ac.uk).

independent feature subsets for utterance-based and segment-based features, respectively. To our knowledge, no such feature selection technique has been applied to segment-based features even in studies based only on a single corpus. Experimental results indicate that the discovered language-independent features yield competitive recognition rates on various corpora in comparison with the full feature set.

The paper is organized as follows. Sect. II presents feature selection strategy. Sect. III overviews all acoustic features relevant to emotions. Sect. IV briefly describes emotional speech corpora employed in this study and presents the experimental settings. Sect. V reports the experimental results and also discusses the selected language-independent features. The last section draws conclusions.

## II. LANGUAGE-INDEPENDENT FEATURES EXPLORATION STRATEGY

The basic idea behind our feature selection strategy is to apply feature selection techniques on a corpus and then test selected features on other corpora of a different language and intonation and repeat this process until language-independent features are found.

Fig. 1 illustrates our proposed strategy with three feature selection techniques for instance. In the strategy, we employ multiple feature selection techniques motivated by the following observations. A single feature selection method often biases to some certain aspects and fails to identify all language-independent features. Thus, the use of selected features results in a lower recognition rate on a different corpus. As three feature selection techniques are used, three feature subsets are generated respectively. As a result, we combine three feature subsets by taking their union or intersection to form two combined subsets of selected features. Then we test combined feature subsets on all the corpora except the one used for feature selection and choose the combined feature set that gives better recognition rates on all the test corpora. The above procedure is repeated by treating each corpus as a training set and testing on all remaining corpora. Finally, feature subsets would be ranked according to recognition rates on majority of the corpora. After the feature subset ranking is obtained, top-ranking subsets are tested with different classifiers based on all corpora to judge their robustness. For stability, all the selected features are also tested on alternative corpora that were not used in feature selection. As a result, the language-independent feature subset is determined via a trade-off between classification performances on all corpora used for feature selection and that of independent corpora. Generally speaking, our proposed feature selection strategy can be viewed as an extended cross-validation procedure where feature selection is made across a set of various corpora. As a result, we summarize this extended cross-validation procedure as a generic algorithm in Fig. 2.

In order to apply the proposed strategy, there are two essential issues worth addressing; i.e., one is the reconciliation of inconsistent emotional states contained in different corpora and the other is the choice of feature selection methods.

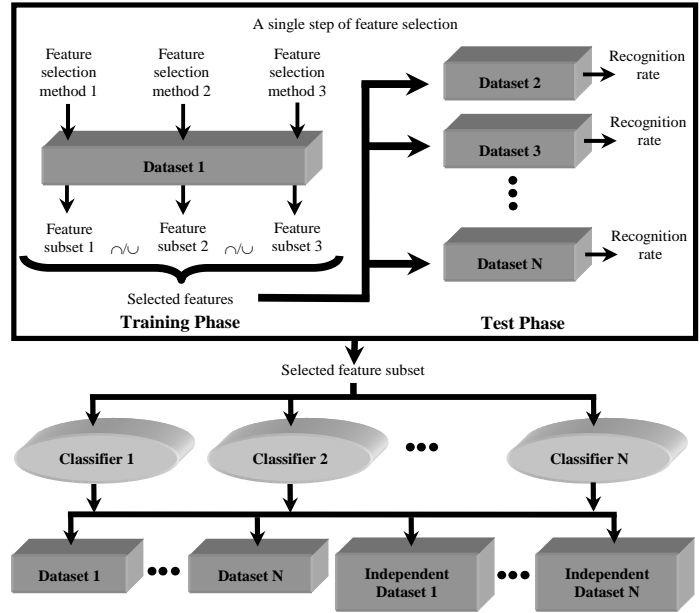

Fig. 1. Schematic illustration of feature selection strategy.

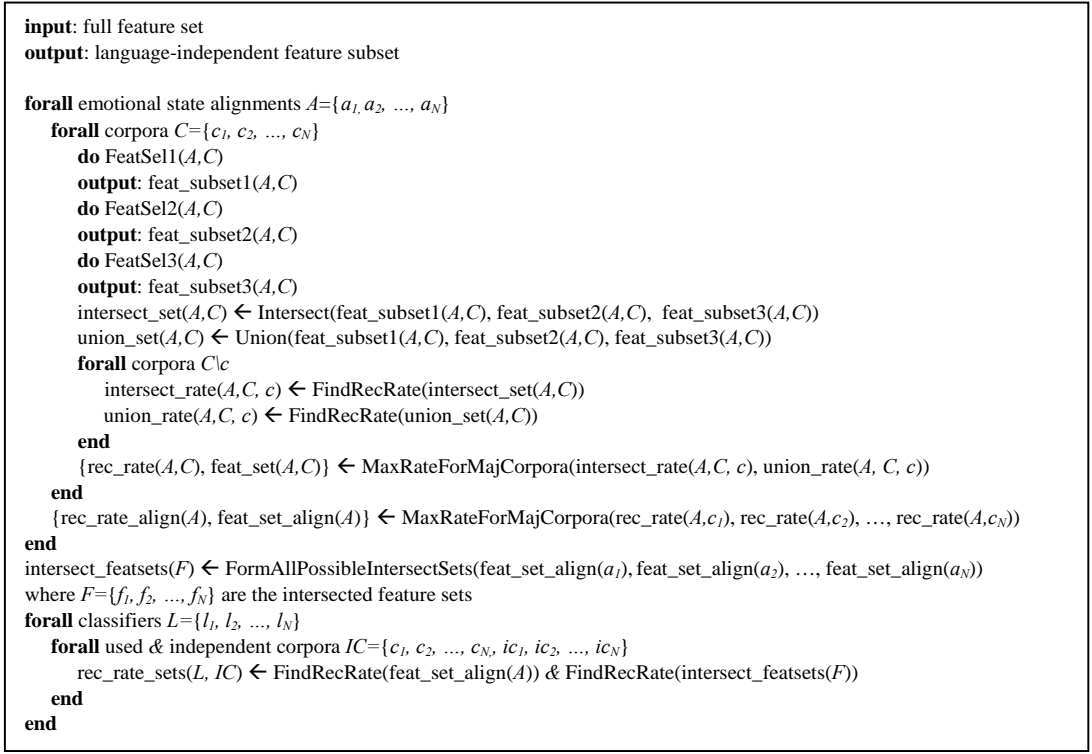

Fig. 2. The feature selection algorithm.

There are often inconsistent emotional states in different emotional speech corpora. In order to discover language-independent features, we do not want them to be dependent on emotional state labels. For this purpose, reconciling emotional states is required in each experiment. A reconciliation scheme may remain a number of emotional states common to all the corpora and/or re-group other emotional states, which results in an *emotional state alignment* so that all corpora have the same emotional states. Based on each alignment, a feature subset is selected by performance evaluation on all test corpora. In this way, the selected language-independent feature subset is stable and tends to yield a high recognition rate for all corpora regardless of emotional states.

In our approach, search strategy and stability are two factors that guide us to choose appropriate feature selection techniques. Different search strategies could result in various feature subsets selected, e.g., local greedy search vs. global search. Stability reflects the sensitivity of selected features to changes in data. Selected features should be robust and lead to good performance regardless of changes in data [8]. Once again, the demand of both diversity of search strategies and the stability justifies the use of multiple feature selection techniques.

### III. ACOUSTIC REPRESENTATION

Acoustic features used for emotional state recognition have been studied from different perspectives. As a result, we collect a set of 318 potentially useful acoustic features reported in literature [1-3, 5, 6, 9, 10]. All 318 features referred to as the *full feature set* hereinafter are grouped in 11 feature categories by their nature as shown in Table I. The joint use of those features in different ways forms two types of representations, *utterance-based* and *segment-based* representations. In general, both representations are required to deal with different types of emotional utterances like words, sentences and passages.

An utterance-based representation treats an utterance as a whole, and hence its representation is formed for the entire utterance. This representation mainly characterizes those global features captured by human listeners. The utterance-based representation used in our work includes all features in the full feature set.

A segment-based representation blocks an utterance into several segments and a feature vector is extracted for each segment. This representation tends to capture critical local features underlying an utterance especially as emotional information is unevenly distributed. The segment-based representation employed in our work has been proposed in [3]. 296 features including the segment length duration and 295 features from other ten categories are used to generate a feature vector for a segment. All feature vectors over segments of an utterance constitute a segment-based representation.

IV. EXPERIMENTAL METHODOLOGY

In this section, we briefly describe emotional speech corpora used in our work and then present our experimental settings.

*A. Emotional Speech Corpora*

In our experiments, we use four speech corpora in four different languages especially designed for emotional speech studies.

The utterances in the Berlin emotional speech corpus are recorded by ten speakers, five actors and five actresses [11]. The corpus consists of total 535 speech files and each speaker uttered ten sentences in German; five are short and five are long sentences. The sentences are daily conversations with no emotional bias [11]. The corpus contains seven emotional states; i.e., happy, sad, anger, neutral, fear/anxiety, boredom and disgust.

In Danish emotional speech (DES) corpus, four speakers, two actors and two actresses, recorded their voices for the corpus [12]. The corpus has 341 emotional utterances in Danish. Unlike the Berlin corpus, an utterance corresponds to one of five emotional states; i.e., happy, sad, anger, neutral and surprise. Every speaker uttered two words, nine sentences and two passages for each emotional state. The script for utterances itself expresses no emotional contents [12].

TABLE I
FULL SET OF VARIOUS ACOUSTIC FEATURES FOR EMOTIONAL SPEECH

| Measure | Full Feature Set |
|---|---|
| Loudness (20) | mean, 25 percentile, 50 percentile, 75 percentile, 25 percentile RMS, 50 percentile RMS, 75 percentile RMS, mean specific loudness band 1 (msl b1), msl b2, msl b3, msl b4, msl b5, msl b6, msl b7, msl b8, msl b9, msl b10, msl b11, msl b12, msl b13 [1]. |
| Voice source (28) | 25 percentile of $E_e$, median of $E_e$, 75 percentile of $E_e$, IQR of normalized $\Delta E_e$, 25 percentile of $\gamma$, median of $\gamma$, 75 percentile of $\gamma$, IQR of normalized $\Delta\gamma$, 25 percentile of $\alpha$, median of $\alpha$, 75 percentile of $\alpha$, IQR of normalized $\Delta\alpha$, 25 percentile of $\beta$, median of $\beta$, 75 percentile of $\beta$, IQR of normalized $\Delta\beta$, 25 percentile of $OQ$, median of $OQ$, 75 percentile of $OQ$, IQR of normalized $\Delta OQ$, 25 percentile of $\varepsilon_o$, median of $\varepsilon_o$, 75 percentile of $\varepsilon_o$, IQR of normalized $\Delta\varepsilon_o$, 25 percentile of $\varepsilon_c$, median of $\varepsilon_c$, 75 percentile of $\varepsilon_c$, IQR of normalized $\Delta\varepsilon_c$ [1]. |
| Other voice source (14) | jitter$_{PF}$, max jitter$_{PQ}$, min jitter$_{PQ}$, shimmer$_{PF}$, max shimmer$_{PQ}$, min shimmer$_{PQ}$, 25 percentile of GNE, median of GNE, 75 percentile of GNE, IQR of normalized $\Delta$GNE, 25 percentile of PSP, median of PSP, 75 percentile of PSP, IQR of normalized $\Delta$PSP [1]. |
| Harmonicity (14) | median of intrinsic diss. $D_I$, range of intrinsic diss. $D_I$, median of avg. diss., median of avg. diss. derivative, median of cons. values at interval $\alpha_1^c$, median of highest cons. interval $\alpha_1^c$, median of cons. values at interval $\alpha_2^c$, median of second highest cons. interval $\alpha_2^c$, median of avg. cons. peak values, median of diss. values at interval $\alpha_1^d$, median of highest diss. interval $\alpha_1^d$, median of diss. values at interval $\alpha_2^d$, median of second highest diss. interval $\alpha_2^d$, median of avg. diss. peak values [1]. |
| Fundamental frequency or pitch (44) | minima series: mean, max, min, range, var, med, 1st quartile, 3rd quartile, iqr, mean abs. val. of derivative. <br> maxima series: mean, max, min, range, var, med, 1st quartile, 3rd quartile, iqr, mean abs. val. of derivative. <br> durations between local extrema series: mean, max, min, range, var, med, 1st quartile, 3rd quartile, iqr, mean abs. val. of derivative. <br> series itself: mean, max, min, range, var, med, 1st quartile, 3rd quartile, iqr, mean abs. val. of derivative, skewness, fraction of voiced F0 above mean, range above mean, range below mean [1], [2]. |
| Intensity or energy (40) | minima series: mean, max, min, range, var, med, 1st quartile, 3rd quartile, iqr, mean abs. val. of derivative. <br> maxima series: mean, max, min, range, var, med, 1st quartile, 3rd quartile, iqr, mean abs. val. of derivative. <br> durations between local extrema series: mean, max, min, range, var, med, 1st quartile, 3rd quartile, iqr, mean abs. val. of derivative. <br> series itself: mean, max, min, range, var, med, 1st quartile, 3rd quartile, iqr, mean abs. val. of derivative [2]. |
| Low-pass intensity (40) | minima series: mean, max, min, range, var, med, 1st quartile, 3rd quartile, iqr, mean abs. val. of derivative. <br> maxima series: mean, max, min, range, var, med, 1st quartile, 3rd quartile, iqr, mean abs. val. of derivative. <br> durations between local extrema series: mean, max, min, range, var, med, 1st quartile, 3rd quartile, iqr, mean abs. val. of derivative. <br> series itself: mean, max, min, range, var, med, 1st quartile, 3rd quartile, iqr, mean abs. val. of derivative [2]. |
| High-pass intensity (40) | minima series: mean, max, min, range, var, med, 1st quartile, 3rd quartile, iqr, mean abs. val. of derivative. <br> maxima series: mean, max, min, range, var, med, 1st quartile, 3rd quartile, iqr, mean abs. val. of derivative. <br> durations between local extrema series: mean, max, min, range, var, med, 1st quartile, 3rd quartile, iqr, mean abs. val. of derivative. <br> series itself: mean, max, min, range, var, med, 1st quartile, 3rd quartile, iqr, mean abs. val. of derivative [2]. |
| Mel-frequency cepstral coefficients (MFCC) (40) | minima series: mean, max, min, range, var, med, 1st quartile, 3rd quartile, iqr, mean abs. val. of derivative. <br> maxima series: mean, max, min, range, var, med, 1st quartile, 3rd quartile, iqr, mean abs. val. of derivative. <br> durations between local extrema series: mean, max, min, range, var, med, 1st quartile, 3rd quartile, iqr, mean abs. val. of derivative. <br> series itself: mean, max, min, range, var, med, 1st quartile, 3rd quartile, iqr, mean abs. val. of derivative [2]. |
| Formant (15) | mean F1, mean F2, mean F3, std F1, std F2, std F3, max F1, max F2, max F3, min F1, min F2, min F3, range F1, range F2, range F3 [5], [9]. |
| Duration (23) | mean dur. of aud. segs., max dur. of aud. segs., min dur. of aud. segs., std. of dur. of aud. segs., mean dur. of inaud. segs., max dur. of inaud. segs., min dur. of inaud. segs., std. of dur. of inaud. segs., no. of aud. segs., no. of inaud. segs., no. of aud. frames., no. of inaud. frames, longest aud. seg., longest inaud. seg., <br> ratios of: no. of aud. to inaud. frames, no. of aud. to inaud. segs., no. of aud. to total no. of frames, no. of aud. to total no. of segs., no. of aud. frames to no. of aud. segs., total duration of aud. segs. to total duration of inaud. segs., duration of aud. segs. to total duration of utterance, duration of inaud. segs. to total duration of utterance, avg. duration of aud. segs. to avg. duration of inaud. segs. [6], [10]. |

The Serbian emotional speech corpus, named GEES, contains 2790 emotional utterances recorded in Serbian [13]. Three actors and three actresses uttered 32 words, 30 short sentences, 30 long sentences and a passage for a single emotional state. The statistics of utterances are phonetically balanced and consistent with the phonetic statistics of Serbian language [13]. The utterances are labeled by five emotional states; i.e., happy, sad, anger, neutral and fear.

The BabyEars emotional speech corpus comprises of natural and spontaneous speech utterances. Unlike the previous corpora, utterances were recorded while six mothers and six fathers naturally talked to their infants [14]. Parents recorded 509 utterances in English language. Sentences and phrases of variable lengths in the corpus contains three emotional states; i.e., approval, attention and prohibition. The emotional states here are totally different from those in three aforementioned corpora.

*B. Experimental Setup*

In our experiments, first three corpora are employed for feature selection while the BabyEars corpus is simply used as a dataset independent of feature selection to monitor the stability of selected feature subsets.

To reconcile different emotional states in the corpora, we design three different reconciliation schemes to produce emotional state alignments. The first scheme adopts the emotional states common to all three corpora only. Thus, four classes; i.e., *happy, sad, anger* and *neutral*, form an emotional state alignment, alignment 1 (A1), so that all utterances corresponding to these four emotional states are used for feature selection. In our second scheme, the emotional states in all the corpora are re-grouped into three categories. For all three corpora, *happy* and *anger* are grouped into the first category. *Sad* and *neutral* are grouped into the second category. The remaining classes in the three corpora are re-labeled as the third category. This reconciliation scheme, alignment 2 (A2), is developed based on the psychological theory that relevant emotion states on a lower level can be re-grouped into a higher taxon in the activation-evaluation space [7]. Similar to the second one, the third scheme named as alignment 3 (A3) adopts four emotional states common to both the DES and the GEES and re-group all other states to form the fifth category. Upon the completion of reconciliation, feature selection is made based on three emotional state alignments, respectively.

For feature selection, we employ three different techniques; i.e., sequential floating forward selection

(SFFS) [15], genetic algorithm (GA) [16] and boosting based feature selection [17]. Basically, these techniques use the wrapper methodology for feature selection. Since our aim is to achieve the maximum recognition rate for emotional state classification with the selected features, a wrapper approach using an error criterion function perfectly meets our requirement. In nature, three techniques adopt different search strategies. The SFFS uses a greedy search strategy and the boosting based method adopts a stage-wise learning style for feature selection whilst the GA based method randomly searches for the best features in a global way. Furthermore, boosting and GA based techniques are quite stable. We anticipate that the joint use of three different feature selection methods leads to a thorough search in the feature space and robustness against substantial changes in corpora. While SFFS and boosting based techniques are applied once to a corpus, the GA based techniques need to run for multiple trials due to its random nature. In our experiments, GA is run for 50 times. The probability of features selected in those runs is calculated and those features of the high probability are selected to form a final feature subset.

The error criterion function used in three feature selection techniques is the K Nearest Neighbor (KNN) [18]. KNN has been widely used for performance evaluation in feature selection due to its advantages, e.g., fast computation, free of parameter tuning etc. [18]. For feature selection, the best value of K is chosen by cross-validation for every corpus on all the alignments. Once a feature subset is obtained, again KNN with 10-fold cross validation is used to achieve a recognition rate for all the test corpora with the selected features. For robustness and efficiency, we verify selected alignment feature subsets by another classifier, Support Vector Machines (SVM) of radial basis function kernel [18] again with 10-fold cross validation for achieving a recognition rate on all corpora.

Once all the feature subsets for the alignments based on the first three corpora are achieved, they will be applied to the BabyEars for a totally different emotional state classification task with the KNN and the SVM working on 10-fold cross-validation. Such results tend to judge how stable and "language-independent" a selected feature subset is. Thus the language-independent feature set is selected based on its performance on both all three corpora and the independent BabyEars corpus.

## V. EXPERIMENTAL RESULTS

In this section, we first report the intermediate classification performance obtained during feature selection with our strategy and then discuss the language-independent acoustic features discovered.

### A. Emotional State Classification

It is observed that feature subsets selected from the Berlin corpus achieve the best testing performance on the DES and the GEES corpora. Thus, the algorithm automatically selects features trained on the Berlin and here we report the results relevant to those feature subsets only. For comparison, we would report all results with both the utterance-based and segment-based *selected feature subsets* (SFS) for three alignments (A1, A2, and A3) and their subset intersection based on different alignments in contrast to the *full feature set* (FFS).

Table II lists the performance produced by KNN on three corpora for those selected utterance-based feature subsets. It is observed that their recognition rates are generally higher than that of the full feature set. As shown in Table III, the performance of the SVM are also comparable to that of the full feature set on those subsets extracted with different alignments but slightly lower on their intersection subsets. Such results suggest that the features selected with the KNN in our approach are robust to a different classifier not used in feature selection and provides evidence for language-independent feature selection.

Moreover, recognition rates on the independent corpus, BabyEars, are also tabulated in Table II and III. It is evident that recognition rates based on all selected feature subset are comparable with that on the full feature set with KNN and so is on some subsets with SVM. Here, we emphasize that the use of selected feature subsets yields the stable yet competitive performance as well even though the BabyEars corpus has totally different emotional states.

TABLE II
RECOGNITION RATE [MEAN (STD)] % OF KNN ON UTTERANCE-BASED FEATURE SUBSETS.

| Corpora | FFS | SFS(A1) | SFS(A2) | SFS(A3) | SFS(A1∩A2) | SFS(A2∩A3) | SFS(A1∩A3) | SFS(A1∩A2∩A3) |
|---|---|---|---|---|---|---|---|---|
| **Features** | 318 | 177 | **161** | 172 | 107 | 130 | 121 | 95 |
| **Berlin** | 66.72 | 65.42 | **68.33** | 68.59 | 66.66 | 68.72 | 68.92 | 66.12 |
|  | (5.00) | (6.76) | **(4.91)** | (6.34) | (5.93) | (5.87) | (5.54) | (7.03) |
| **DES** | 54.82 | 58.98 | **56.92** | 56.91 | 60.18 | 58.08 | 58.38 | 57.51 |
|  | (8.44) | (8.50) | **(9.80)** | (8.27) | (6.95) | (9.44) | (7.60) | (8.27) |
| **GEES** | 72.87 | 73.19 | **76.46** | 75.13 | 75.70 | 75.81 | 74.63 | 75.35 |
|  | (2.67) | (2.03) | **(3.20)** | (3.03) | (2.15) | (2.31) | (2.06) | (3.10) |
| **BabyEars** | 57.94 | 61.10 | **57.55** | 59.13 | 57.36 | 59.71 | 58.93 | 57.55 |
|  | (7.77) | (6.53) | **(7.49)** | (6.88) | (8.93) | (8.01) | (9.45) | (8.16) |

TABLE III
RECOGNITION RATE [MEAN (STD)] % OF SVM ON UTTERANCE-BASED FEATURE SUBSETS.

| Corpora | FFS | SFS(A1) | SFS(A2) | SFS(A3) | SFS(A1∩A2) | SFS(A2∩A3) | SFS(A1∩A3) | SFS(A1∩A2∩A3) |
|---|---|---|---|---|---|---|---|---|
| **Features** | 318 | 177 | **161** | 172 | 107 | 130 | 121 | 95 |
| **Berlin** | 80.53 | 80.00 | 82.05 | 80.18 | 79.03 | 82.05 | 79.24 | 78.46 |
| | (5.20) | (5.78) | **(3.81)** | (3.91) | (3.54) | (4.70) | (3.99) | (3.60) |
| **DES** | 71.67 | 70.48 | 72.58 | 71.11 | 68.46 | 70.54 | 70.51 | 68.44 |
| | (7.93) | (8.63) | **(7.44)** | (6.68) | (7.39) | (8.00) | (8.17) | (8.76) |
| **GEES** | 87.08 | 85.65 | 86.87 | 85.61 | 85.65 | 85.94 | 84.86 | 84.21 |
| | (1.73) | (1.88) | **(1.78)** | (2.15) | (1.85) | (2.04) | (1.92) | (1.62) |
| **BabyEars** | 64.63 | 62.85 | 63.65 | 61.88 | 62.85 | 61.28 | 62.85 | 61.67 |
| | (5.46) | (6.30) | **(7.95)** | (6.17) | (7.28) | (6.58) | (8.72) | (7.17) |

TABLE IV
RECOGNITION RATE [MEAN (STD)] % OF KNN ON SEGMENT-BASED FEATURE SUBSETS.

| Corpora | FFS | SFS(A1) | SFS(A2) | SFS(A3) | SFS(A1∩A2) | SFS(A2∩A3) | SFS(A1∩A3) | SFS(A1∩A2∩A3) |
|---|---|---|---|---|---|---|---|---|
| **Features** | 296 | **125** | 180 | 152 | 94 | 128 | 89 | 80 |
| **Berlin** | 62.83 | 65.33 | 65.29 | 64.14 | 63.38 | 64.16 | 64.31 | 64.81 |
| | (6.75) | **(9.35)** | (5.35) | (7.29) | (5.52) | (6.01) | (5.80) | (7.55) |
| **DES** | 51.19 | 51.21 | 50.90 | 49.14 | 49.71 | 47.10 | 48.55 | 48.84 |
| | (7.92) | **(6.42)** | (4.09) | (6.20) | (5.52) | (5.84) | (6.47) | (5.48) |
| **GEES** | 75.91 | 76.44 | 74.72 | 76.01 | 73.92 | 75.30 | 74.39 | 73.60 |
| | (2.55) | **(1.87)** | (2.11) | (2.54) | (2.26) | (3.24) | (2.56) | (2.24) |
| **BabyEars** | 56.80 | 56.00 | 57.58 | 56.00 | 55.80 | 56.40 | 55.80 | 55.00 |
| | (8.49) | **(10.29)** | (10.15) | (9.04) | (9.44) | (9.21) | (10.32) | (9.41) |

TABLE V
RECOGNITION RATE [MEAN (STD)] % OF SVM ON SEGMENT-BASED FEATURE SUBSETS.

| Corpora | FFS | SFS(A1) | SFS(A2) | SFS(A3) | SFS(A1∩A2) | SFS(A2∩A3) | SFS(A1∩A3) | SFS(A1∩A2∩A3) |
|---|---|---|---|---|---|---|---|---|
| **Features** | 296 | **125** | 180 | 152 | 94 | 128 | 89 | 80 |
| **Berlin** | 74.59 | 73.33 | 74.38 | 74.24 | 73.16 | 73.68 | 73.72 | 72.42 |
| | (6.64) | **(6.06)** | (4.27) | (5.33) | (5.77) | (5.49) | (6.30) | (8.46) |
| **DES** | 57.15 | 55.96 | 57.13 | 58.03 | 56.25 | 56.86 | 55.66 | 55.96 |
| | (8.33) | **(8.15)** | (7.29) | (8.75) | (8.54) | (9.70) | (8.55) | (8.38) |
| **GEES** | 85.16 | 83.47 | 83.51 | 84.19 | 82.86 | 83.32 | 83.11 | 82.06 |
| | (2.40) | **(2.91)** | (2.07) | (2.29) | (2.96) | (1.67) | (2.25) | (2.36) |
| **BabyEars** | 62.88 | 59.95 | 58.96 | 58.56 | 60.72 | 58.17 | 59.73 | 60.13 |
| | (10.14) | **(10.52)** | (8.21) | (8.49) | (6.79) | (7.24) | (6.30) | (7.38) |

Thus, our feature selection algorithm chooses a feature subset which yields the maximum recognition rate on all four corpora. In terms of KNN, the nearly same performance is achieved on different feature subsets. For SVM, however, the feature subset extracted with alignment 2 leads to the best performance on all the corpora. Thus, this feature subset of 161 features is chosen as language-independent feature set as is stable and robust against substantial changes on different corpora.

Likewise, feature selection on segment-based features is performed to select a language-independent feature subset for this representation. Table IV shows the KNN results produced during feature selection. With the KNN, the recognition rates on feature subsets are almost equal to that on the full feature set.

Similarly, the recognition rates with SVM are reported in Table V. By the same trade-off as done for the utterance-based representation, we pick the feature subset of 125 features extracted with alignment 1 to be the language-independent feature set for the segment-based representation since overall this subset leads to the best generalization performance.

*B. Language-independent Acoustic Features*

The selected utterance-based and segment-based language-independent feature subsets are collectively shown in Table VI. As observed from Table VI, there are 87 acoustic features in ten categories shared between two language-independent feature subset, highlighted in bold. The fact suggests that language-independent features selected in terms of two different acoustic representations are mostly common. In other words, those common features are essential to characterize emotional speech at different levels of speech communication.

Loudness features refer to the degree of audibility in different sounds [1]. As a result, 15 and 19 features are selected for utterance-based and segment-based representation, respectively, which suggests the importance of loudness features in characterizing emotional speech. Features relevant to voice source encode the perceptual effects on variations in the voice source or glottal excitation [1]. 18 and 17 features derived from Liljencrants-Fant parameterization of the glottal volume velocity signal are selected for utterance-based and segment-based representations, respectively. Other voice source features reflect variations in pitch, energy, glottal flow spectral decay and excitation noise [1]. As a result, 12 features are picked for the utterance-based representation but only 4 are selected for the segment-based representation. It suggests that such features are more important to characterize emotions at the utterance level. It is known that harmonic features directly relate to patterns of spectral harmonics in the voiced part of speech [1]. 11 and seven spectral harmonic features are selected for utterance-based and segment-based representations, respectively. In [1], the SFFS method was applied to a subset of our FSS based on a single corpus for feature selection. By comparison, we find that most of the loudness, voice source, other voice source, and harmonicity features selected using our method are also on their top 50 feature list [1], while other selected features mentioned above are not in their selected feature subset as they used a data-dependent feature selection method with a single corpus.

TABLE VI
UTTERANCE-BASED AND SEGMENT-BASED LANGUAGE-INDEPENDENT FEATURES

| Measure | Utterance-based Language-independent Features | Segment-based Language-independent Features |
|---|---|---|
| Loudness | **mean, 25 percentile, 50 percentile, 75 percentile, 75 percentile RMS, msl b4, msl b5, msl b6, msl b7, msl b8, msl b9, msl b10, msl b11, msl b12, msl b13**. | **mean, 25 percentile, 50 percentile, 75 percentile**, 25 percentile RMS, 50 percentile RMS, **75 percentile RMS,** msl b1, msl b3, **msl b4, msl b5, msl b6, msl b7, msl b8, msl b9, msl b10, msl b11, msl b12, msl b13**. |
| Voice source | **median of $E_e$, 75 percentile of $E_e$, IQR of normalized $\Delta E_e$, 75 percentile of $\gamma$, 25 percentile of $\alpha$, 75 percentile of $\alpha$,** IQR of normalized $\Delta\alpha$, **25 percentile of $\beta$, median of $\beta$, 75 percentile of $\beta$, IQR of normalized $\Delta OQ$,** 25 percentile of $\varepsilon_o$, **median of $\varepsilon_o$,** 75 percentile of $\varepsilon_o$, 25 percentile of $\varepsilon_c$, **median of $\varepsilon_c$,** 75 percentile of $\varepsilon_c$, IQR of normalized $\Delta\varepsilon_c$. | 25 percentile of $E_e$, **median of $E_e$, 75 percentile of $E_e$, 75 percentile of $\gamma$, 25 percentile of $\alpha$,** median of $\alpha$, **75 percentile of $\alpha$, 25 percentile of $\beta$, median of $\beta$, 75 percentile of $\beta$,** IQR of normalized $\Delta\beta$, median of $OQ$, **IQR of normalized $\Delta OQ$, median of $\varepsilon_o$, 75 percentile of $\varepsilon_o$,** IQR of normalized $\Delta\varepsilon_o$, **median of $\varepsilon_c$**. |
| Other voice source | jitter$_{PF}$, max jitter$_{PQ}$, min jitter$_{PQ}$, **max shimmer$_{PQ}$,** min shimmer$_{PQ}$, 25 percentile of GNE, **median of GNE,** 75 percentile of GNE, 25 percentile of PSP, **median of PSP,** 75 percentile of PSP, **IQR of normalized $\Delta$PSP**. | **max shimmer$_{PQ}$, median of GNE, median of PSP, IQR of normalized $\Delta$PSP**. |
| Harmonicity | **median of intrinsic diss. $D_I$,** range of intrinsic diss. $D_I$, **median of avg. diss., median of avg. diss. derivative, median of cons. values at interval $\alpha_1^c$, median of highest cons. interval $\alpha_1^c$, median of cons. values at interval $\alpha_2^c$, median of avg. cons. peak values,** median of diss. values at interval $\alpha_1^d$, median of diss. values at interval $\alpha_2^d$, median of avg. diss. peak values. | **median of intrinsic diss. $D_I$, median of avg. diss., median of avg. diss. derivative, median of cons. values at interval $\alpha_1^c$, median of highest cons. interval $\alpha_1^c$, median of cons. values at interval $\alpha_2^c$, median of avg. cons. peak values**. |
| Fundamental frequency or pitch | minima series: mean, **max, min, range, med,** 1st quartile, 3rd quartile, iqr.<br>maxima series: range, **med,** 1st quartile, 3rd quartile.<br>durations between local extrema series: **min,** range, **med**.<br>series itself: mean, max, **min, range, var, med,** 1st quartile, **3rd quartile, iqr, mean abs. val. of derivative, skewness, fraction of voiced F0 above mean,** range above mean, **range below mean.** | minima series: **max, min, range,** var, **med,** mean abs. val. of derivative.<br>maxima series: min, var, **med, 1st quartile,** iqr.<br>durations between local extrema series: **min, med,** 1st quartile.<br>series itself: **min, range, var, med,** 3rd quartile, iqr, mean abs. val. of derivative, skewness, fraction of voiced F0 above mean, range below mean. |
| Intensity or energy | minima series: min, med, 1st quartile, iqr.<br>maxima series: mean abs. val. of derivative.<br>series itself: min, var. | minima series: var.<br>durations between local extrema series: var. |
| Low-pass intensity | minima series: **mean, max, range, var, med,** 1st quartile, **3rd quartile, iqr, mean abs. val. of derivative.**<br>maxima series: **mean, max, var, med, 3rd quartile,** iqr, **mean abs. val. of derivative.**<br>durations between local extrema series: max, **min,** range, var, 3rd quartile, iqr, **mean abs. val. of derivative.**<br>series itself: **mean, max,** min, **var, med, 3rd quartile, iqr, mean abs. val. of derivative**. | minima series: **mean, max,** min, **range, var,** med, 3rd quartile, iqr, mean abs. val. of derivative.<br>maxima series: **mean, max,** min, range, **var, med,** 1st quartile, **3rd quartile, mean abs. val. of derivative.**<br>durations between local extrema series: mean, **min,** med, **mean abs. val. of derivative.**<br>series itself: **mean, max,** range, **var, med,** 1st quartile, **3rd quartile, iqr, mean abs. val. of derivative**. |
| High-pass intensity | minima series: min, 1st quartile.<br>maxima series: max, min, range, med, 1st quartile, mean abs. val. of derivative.<br>series itself: mean, max, **min**. | minima series: var, med.<br>series itself: **min**. |
| Mel-frequency cepstral coefficients (MFCC) | minima series: mean, **med, 1st quartile,** 3rd quartile, **mean abs. val. of derivative.**<br>maxima series: min, **1st quartile,** iqr, mean abs. val. of derivative.<br>durations between local extrema series: var, med, 1st quartile.<br>series itself: mean, **med,** 3rd quartile, iqr, mean abs. val. of derivative. | minima series: max, range, **med, 1st quartile, mean abs. val. of derivative.**<br>maxima series: range, **1st quartile.**<br>durations between local extrema series: 3rd quartile.<br>series itself: max, range, **med,** 1st quartile. |
| Formant | **mean F1, mean F2,** mean F3, std F2, std F3. | **mean F1, mean F2,** std F1, max F1, min F1, range F1. |
| Duration | mean dur. of aud. segs., min dur. of aud. segs., std. of dur. of aud. segs., ratios of: no. of aud. to total no. of frames, duration of aud. segs. to total duration of utterance. | N/A |

On the other hand, the pitch related features are mainly the statistics of minima pitch series and original pitch series, which is likely to reflect emotional variation in speech. As a result, 29 and 24 pitch related features are selected for utterance-based and segment-based representations. Our results here lend a support to previous studies on the relationship between pitch and emotion [2], [19]. Regarding intensity related features, the majority selected are due to the low-pass intensity measure out of the three intensity measures. 31 statistical low-pass intensity features are selected for utterance-based and segment-based representations, respectively, which portrays the significance of these features. In contrast, only seven utterance-based and two segment-based features are selected from intensity measure. Regarding features achieved with the high-pass intensity measure, 11 utterance-based features and three segment-based features only are selected. For different purposes in speech information processing, Mel-frequency cepstral coefficients (MFCC) are extensively used to characterize speech. In our experiments, we find out 17 and 12 MFCC-based features likely to be responsible for emotional speech for utterance-based and segment-based representations, respectively. Pitch and the intensity based features were discussed in terms of emotion [2], where the importance of such features was ranked with information gain. Top 20 features listed in [2] are six low-pass intensity and 14 pitch related features, and most of them are also selected as language-independent features by our method. Nevertheless, none of the MFCC-related features was on their top feature list, while ours selects such features as language-independent features. We believe that they could be useful by combining other emotional acoustic features though they do not have a very high information gain as used individually.

Finally, five and six formant related features are picked for utterance-based and segment-based representations, respectively. The selected features are the $1^{st}$ and the $2^{nd}$ order statistics of formants. Feature selection results in [5] with the SFFS on a single corpus also include two formant features in their top ten feature lists. In addition, our method selects five duration features for the utterance-based representation. Only three duration features are selected in [6] where the SFFS was used on a single corpus. Our results suggest that duration features may be salient for some spoken languages but less important in characterizing emotional speech language-independently.

VI. CONCLUSION

We have presented a novel feature selection strategy by using multiple feature selection methods on different emotional speech corpora. Based on a number of corpora available, we identify two language-independent feature subsets, respectively, for utterance-based and segment-based representations. As we have only four corpora at hand, the results reported here simply reflect our efforts to uncover language-independent emotional acoustic features. These selected features would be verified further when more corpora are available. On the other hand, listening tests on four corpora we used were all performed by native listeners. Thus, our results reported in Sect. V.A also provide the baseline performance of automatic classification for further listening tests conducted in the future psychological studies, e.g., whether emotion over speech can be detected and recognized by listeners who are not native speaker without understanding the linguistic content of an utterance. We firmly believe that the exploration of language-independent emotional acoustic features is of significance for all kinds of speech information processing tasks ranging from speech synthesis to speech recognition.

ACKNOWLEDGMENT

Authors are grateful to the owners of the Berlin, DES, GEES and BabyEars emotional speech corpora for providing their corpora. Authors are also grateful to R. Fernandez for providing his feature extraction code. A. Shaukat would like to thank National University of Sciences and Technology, Pakistan for their financial support.